\documentclass[10pt,twocolumn,letterpaper]{article}

\usepackage{cvpr}              

\usepackage{graphicx}
\usepackage{amsmath}
\usepackage{amssymb}
\usepackage{booktabs}
\usepackage{enumitem}

\usepackage{bm}
\usepackage{amsmath}
\usepackage{colortbl}
\usepackage[dvipsnames,svgnames]{xcolor}
\usepackage{blindtext}
\usepackage{lipsum,lineno}
\usepackage{multirow}
\usepackage{tikz}
\usepackage{wrapfig}
\usepackage{cases}
\usepackage{pifont}
\newcommand{\mytilde}{\raise.17ex\hbox{$\scriptstyle\mathtt{\sim}$}}
\usepackage[pagebackref,breaklinks,colorlinks,bookmarks=false]{hyperref}

\usepackage[capitalize]{cleveref}
\crefname{section}{Sec.}{Secs.}
\Crefname{section}{Section}{Sections}
\Crefname{table}{Table}{Tables}
\crefname{table}{Tab.}{Tabs.}

\newcommand\blfootnote[1]{%
  \begingroup
  \renewcommand\thefootnote{}\footnote{#1}%
  \addtocounter{footnote}{-1}%
  \endgroup
}

\def\tabref#1{Tab.~\ref{#1}}
\def\figref#1{Fig.~\ref{#1}}

\def\cvprPaperID{3790} 
\def\confName{CVPR}
\def\confYear{2024}

\definecolor{peach}{rgb}{ 0.943, 0.188, 0.526}
\definecolor{plum}{rgb}{ 0.858, 0.188, 0.478}
\definecolor{muted_navy_blue}{RGB}{63, 75, 166}
\definecolor{muted_sky_blue}{RGB}{134,166,213}
\definecolor{federal_blue}{RGB}{0,96,240}
\definecolor{regulation_red}{RGB}{226, 20, 79}
\definecolor{federal_gold}{RGB}{240, 212, 14}

\newcommand{\rev}[1]{{#1}}

\newcommand{\papername}{DiffAvatar}%

\begin{document}

\title{\papername : Simulation-Ready Garment Optimization \\ with Differentiable Simulation}  

%

\author{Yifei Li$^{*2}$ \hspace{0.3em} Hsiao-yu Chen$^{1}$ \hspace{0.3em} Egor Larionov$^{1}$ \hspace{0.3em} Nikolaos Sarafianos$^{1}$ \hspace{0.3em} Wojciech Matusik$^2$ \hspace{0.3em} Tuur Stuyck$^{1}$ \\
{\normalsize $^1$Meta Reality Labs} \quad
{\normalsize $^2$MIT CSAIL} 
}


\twocolumn[{%
\renewcommand\twocolumn[1][]{#1}%
\maketitle
\vspace{-1.0cm}
\centering
    \includegraphics[width=0.99\textwidth]{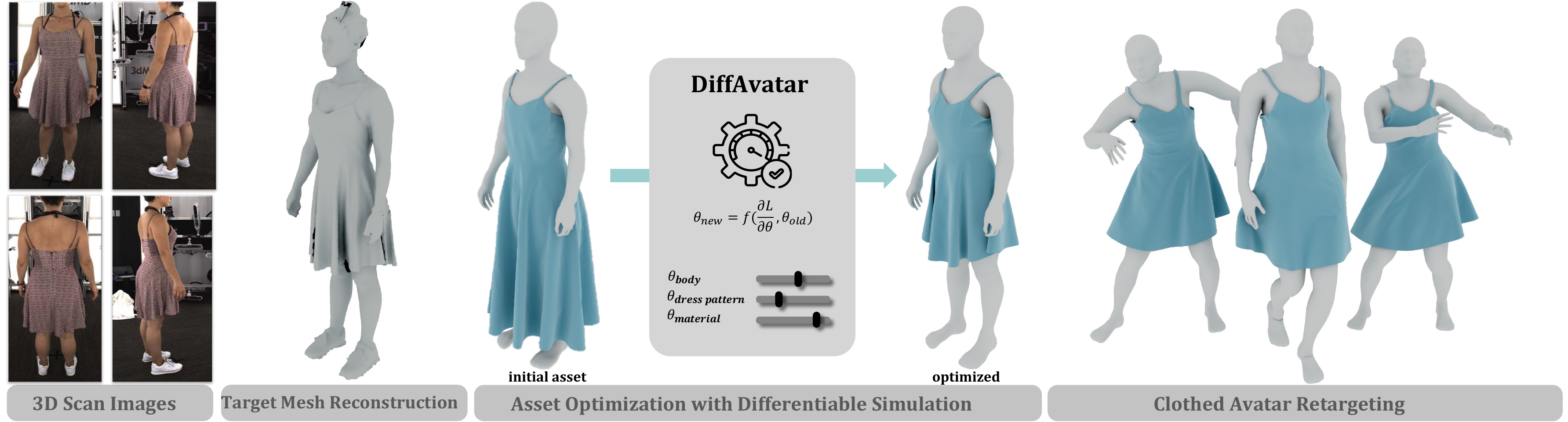}
    \captionof{figure}{
        \textit{We present \textbf{\papername}, an automated computational method to recover simulation-ready garment and body assets. Starting from a multi-view capture, we reconstruct a semantically segmented 3D mesh. The segmented clothing geometry acts as a target shape for our optimization pipeline. Our method recovers body shape and pose, clothing pattern and clothing material parameters from a single scan. We optimize a clothing template in 2D pattern space to reproduce the captured clothing in 3D in a physical way. We compute gradients of required parameters using a differentiable simulation approach.}
    }
    \vspace*{0.2cm}
    \label{fig:teaser}
}]

\begin{abstract}
\vspace{-0.3cm}
The realism of digital avatars is crucial in enabling telepresence applications with self-expression and customization. While physical simulations can produce realistic motions for clothed humans, they require high-quality garment assets with associated physical parameters for cloth simulations. 
However, manually creating these assets and calibrating their parameters is labor-intensive and requires specialized expertise. \rev{Current methods focus on reconstructing geometry, but don't generate complete assets for physics-based applications.}
To address this gap, we propose \papername,~a novel approach that performs body and garment co-optimization using differentiable simulation. By integrating physical simulation into the optimization loop and accounting for the complex nonlinear behavior of cloth and its intricate interaction with the body, our framework recovers body and garment geometry and extracts important material parameters in a physically plausible way. \blfootnote{*This work was conducted during an internship at Meta Reality Labs}
Our experiments demonstrate that our approach generates realistic clothing and body shape suitable for downstream applications. We provide additional insights and results on our webpage: \href{https://people.csail.mit.edu/liyifei/publication/diffavatar/}{people.csail.mit.edu/liyifei/publication/diffavatar}.
\end{abstract}


\section{Introduction} \label{sec:intro}
Virtual avatars are increasingly gaining importance as they serve as a digital extensions of users, enabling novel social and professional interactions. The physical realism of avatars, including realistic clothing and accurate body shape, is crucial for such applications. This need for realism extends beyond visual aesthetics but also includes dynamic interactions and motion obtained by accurate physical simulation of clothing and body dynamics. Physical simulation and rendering techniques can be used as tools to achieve physical realism in the virtual world. However, this requires the creation of high-quality clothing assets for individual users, which presents a substantial challenge. The conventional approach requires meticulous manual design by artists, a process that is exceedingly time-consuming. 
This manual approach is fundamentally unfeasible for individualized avatar clothing, especially considering the continuously growing user base of telepresence applications. 
The notion of having an artist create a unique virtual outfit for every user is simply impractical. 
This scenario underscores the pressing need for automated solutions for scalable and personalized avatar asset creation and optimization. 
Recent advancements in computer vision and graphics have accelerated the automation of avatar asset creation from user images or scans. However, the predominant focus has been on geometry reconstruction, with limited focus on generating complete assets that can be used in physics-based applications. 

\papername~endeavors to bridge this gap by introducing a body and garment co-optimization pipeline using differentiable simulation. By entwining physical simulation within the optimization loop, we ensure that the dynamics of the clothing are considered in the optimization process. \rev{We optimize for all assets required for physics-based simulation and other downstream applications in a physically plausible way by leveraging differentiable cloth simulation for body shape recovery and extending it to optimize for garment shape directly in the rest shape pattern space. Specifically,} we recover garment patterns, body pose and shape, as well as retrieving the crucial physical material parameters leveraging only a minimal garment template library. 
We believe that our work is the first to leverage high-resolution differentiable simulation for asset recovery from real scans which often contain holes and compromised boundaries. In summary, our key contributions are as follows:
\begin{itemize}[leftmargin=*]
\itemsep-0.1em 
    \item
        A novel approach that utilizes differentiable simulation for co-optimizing garment shape and materials, and body shape and pose, while taking into account cloth deformations and collisions in the context of avatar asset recovery.
    \item 
        A unified method for body shape, pose and garment assets recovery from one \textit{real noisy 3D scan of a clothed person.} 
    \item 
        \rev{For the first time in a differentiable cloth simulation algorithm, we incorporate optimization through the cloth rest shape. Additionally, we develop a differentiable control cage representation for garment shape optimization to regularize the 2D garment pattern space and produce effective optimization results.} 
\end{itemize}

\section{Related Work}

\begin{figure*}[t]
    \centering
    \includegraphics[width=1.0\textwidth]{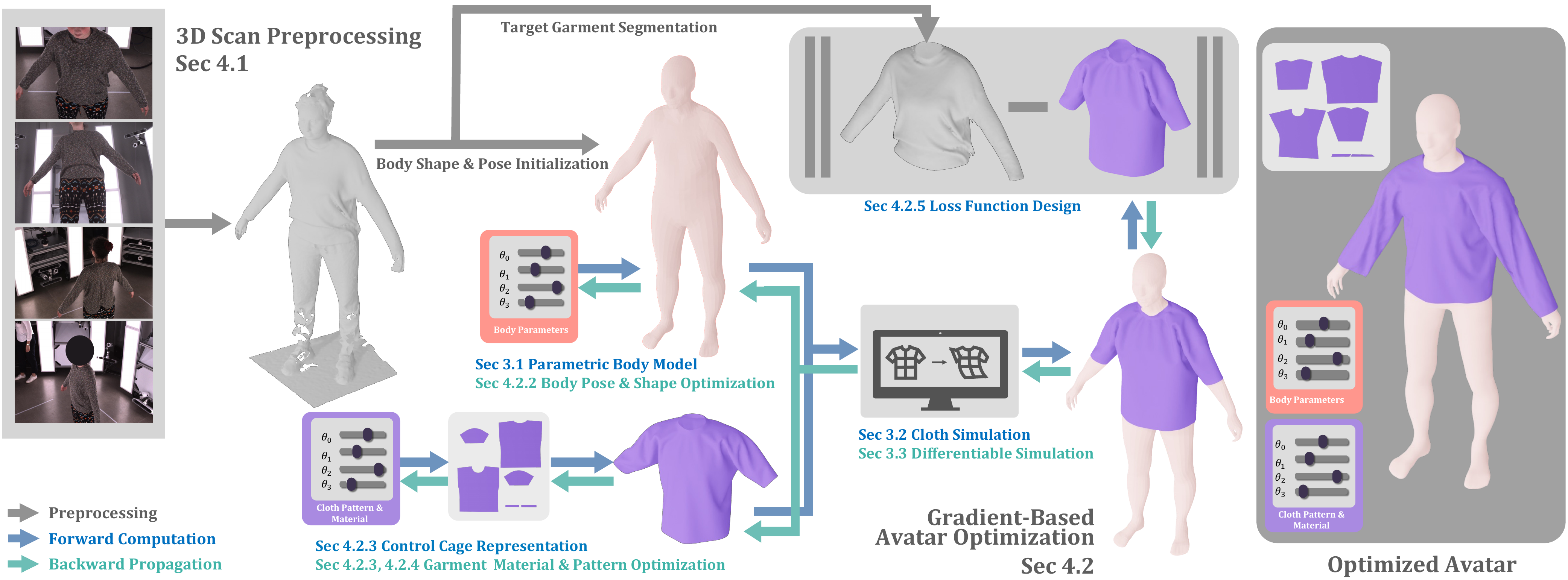}
    \caption{\textbf{\papername~} generates simulation-ready avatar assets from inputs obtained through a multi-view capture. Our pipeline initially preprocesses the 3D scan to segment the target garment and establish the initial pose and shape of the parametric body model. We employ a differentiable simulation framework to align our simulated garment with the segmented garment by jointly optimizing the garment's design and material parameters, along with the body shape.}
    \label{fig:overview}
    \vspace{-0.3cm}
\end{figure*}

\noindent\textbf{Pose and Shape Estimation} precedes the garment shape estimation and its properties since the underlying body directly impacts how cloth drapes and behaves when in motion. 
Prior works focus on reconstructing the body from minimally clothed~\cite{anguelov2005, SMPL_model} captures or focus on complex captures of clothed humans~\cite{Yang2018, li2021deep, Bang2021, chen2021tightcap, liang2022fabric} to extract a representation for body and garments. 
These methods build on SMPL~\cite{SMPL_model}, and work in conjunction with simulated \cite{Bang2021, li2021deep, liang2022fabric, Yang2018}, or trained \cite{chen2021tightcap} garment models.

\noindent\textbf{Cloth Simulation} methods have been widely used for digitally modeling the behavior of fabrics in visual effects and movie productions since the pioneering work on implicit cloth simulation~\cite{baraff1998large} and follow-up works~\cite{choi2005stable, grinspun2003discrete, gingold2004discrete} allowing for stable and efficient simulations. Position Based Dynamics~(PBD) \cite{muller2007position} updates positions directly to project constraints in a highly parallel fashion resulting in high performance simulations. eXtended Position Based Dynamics (XPBD)~\cite{macklin2016xpbd} overcome the limitation of iteration dependent behavior of PBD while Projective Dynamics~\cite{bouaziz2014projective} connects nodal Finite Element methods and Position Based methods, leading to an efficient and accurate solver. Stuyck~\cite{stuyck2022cloth} provides an overview on cloth simulation techniques.

\noindent\textbf{Garment Pattern Estimation} is essential as the 2D sewing pattern influences the fit and the formation of wrinkles on the 3D body. One approach involves flattening the 3D shape into several developable~\cite{stein2018developability} 2D pieces. However, these methods require manual cutting input to generate pieces with minimal distortion~\cite{Bang2021, Pietroni2022}. Other works use neural networks to learn the seams \cite{Goto2021DatadrivenGP}, but the patterns obtained through direct flattening of the 3D shape are only suitable for nearly undeformed cloth, which is rarely the case in real-world draped garments. 
Follow-up works employed neural networks to learn the 2D rest shape and yield more accurate patterns, but their generality is limited to the training data and specific garments~\cite{Yang2018, ChenNeuralSewing2022}. Parameterized garment patterns~\cite{Korosteleva2022, GarmentCode2023} address these limitations and can adapt to a wide range of shapes but struggle to generalize to real garments and lack control over symmetry and matching seam lines.
Alternatively 2D patterns can be optimized using a physics simulator in an iterative manner~\cite{WangRuleFreeSewing2018, Bartle2016, Wolff2023}.

\noindent\textbf{Differentiable Simulation} allows for gradient computation with respect to simulation parameters, enabling the use of gradient-based optimization algorithms to find solutions for inverse design and system identification. Early works applied the adjoint method to fluid~\cite{mcnamara2004fluid, li2022anisotropicStokes} and cloth~\cite{wojtan2006keyframe} simulation models to analytically compute gradients. Recent techniques differentiate through complex simulations such as Projective Dynamics \cite{du2021diffpd} and XPBD \cite{stuyck2023diffxpbd}. Differentiable simulation methods have successfully been applied to cloth simulation with frictional contact \cite{li2022diffcloth}, material estimation \cite{chen2022virtual, Larionov} and shape and pose estimation \cite{guo2021inverse}.

\noindent\textbf{Learning-based approaches} have focused on garment draping, modeling of cloth dynamics, or handling collisions and contact~\cite{rodriguez2023will, kairanda2023neuralclothsim, casado2022pergamo, bertiche2022neural, li2022dig, grigorev2023hood, santesteban2022snug, tiwari21neuralgif, xue2023nsf, romero2022contact, santesteban2021self, su2023caphy, lee2023multi, liu2023sewformer, halimi2022pattern, xiang2022dressing} and hair \cite{wang2022hvh, Wang_2023_CVPR} and the introduction of new large-scale datasets~\cite{zou2023cloth4d} albeit synthetic, will further accelerate this progress. DrapeNet~\cite{de2023drapenet} predicts a 3D deformation field conditioned on the latent codes of a generative network, which models garments as unsigned distance fields allowing it to handle and edit unseen clothes. Qiu \etal~\cite{qiu2023rec} reconstruct 3D clothes from monocular videos using SDFs and deformation fields. Qi~\etal~\cite{qi2023personaltailor} proposed a personalized 2D pattern design method using synthetic data, where the user can input specific constraints for personal 2D pattern design from 3D point clouds. 
%
Li~\etal\cite{li2023isp} proposed a parametric garment representation model for garment draping using SDFs. 
\vspace{-0.3cm}
\section{Preliminaries}
\subsection{Body and Garment Models}
\label{eq:PCA}

\noindent\textbf{Body Shape and Pose} are represented using a parameterized statistical body model similar to SMPL~\cite{SMPL_model} in our method. The skeleton is defined by joints which are described by $P$ parameters encoding local transformations through joint angles $\psi$ and bone lengths.  The shape is encoded by the statistical shape coefficients $\bm{\nu}$ as $\mathcal{V}_0 + \bm{\nu} \mathcal{V}$
where $\mathcal{V}_0$ and $\mathcal{V}$ encode the average body shape and the shape basis functions respectively. The body shape with $V_b$ vertices is posed with the skeleton using a linear blend skinning function $\mathcal{S} : \mathbb{R}^{3\times V_b} \times \mathbb{R}^{P} \rightarrow \mathbb{R}^{3\times V_b}$ \cite{magnenat1988joint}.

\begin{figure}[t]
    \centering
    \includegraphics[width=0.49\textwidth, trim={0 0 0 0},clip]{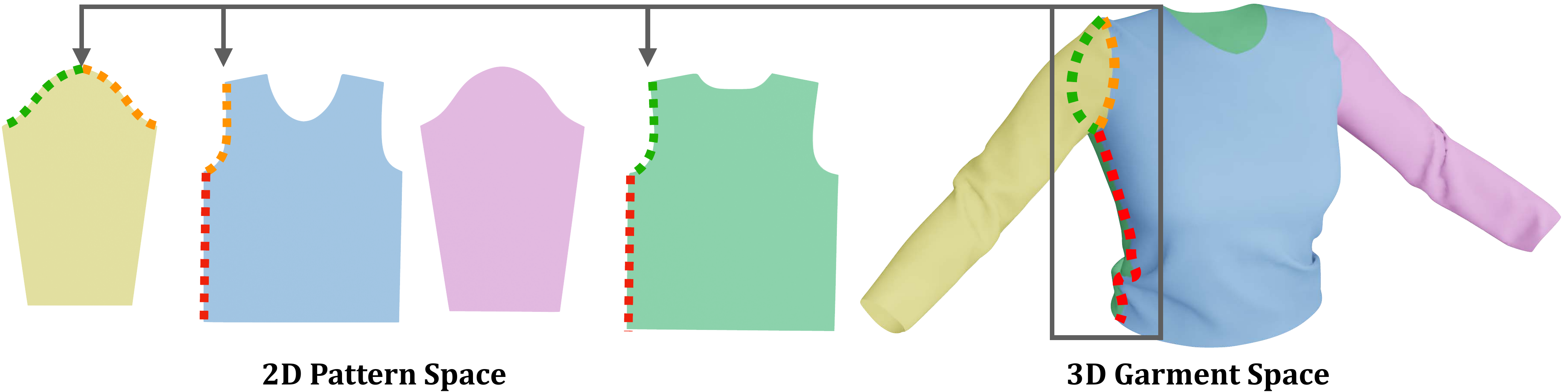}
    \caption{3D garments (right) can be compacted represented as their 2D panels (left). Seams are visualized as dotted-lines.  }
    \label{fig:patternspaces}
    \vspace{-0.3cm}
\end{figure}

\noindent\textbf{Garments} can take on a wide range of 3D shapes when draped onto a body, due to factors such as changing pose and dynamics or wearer manipulations. Despite this large variation in configurations, garments are compactly represented by their 2D patterns (Fig.~\ref{fig:patternspaces}), which consist of the individual pieces of fabric that are sewn together to create the 3D clothing. 
Therefore, we represent clothing in 2D pattern space, which ensures developable~\cite{stein2018developability} meshes and manufacturable clothing. 
Virtual garments are modeled as triangle meshes, with their rest shape encoded in these 2D patterns. The rest shape is crucial for modeling the in-plane stretching and shearing behavior of different fabrics.

\subsection{Cloth Simulation}

We compute the deformation of a garment mesh consisting of $V$ vertices which is draped on a posed body using dynamic physics-based simulation. The simulator effectively solves Newton's equations of motion given by $\mathbf{M}\dot{\mathbf{v}} = -\nabla U(\mathbf{x}),$ where $\mathbf{x} \in \mathbb{R}^{3V}$ and $\mathbf{v} \in \mathbb{R}^{3V}$ are the vertex positions and velocities, $U(\mathbf{x})$ is the energy potential and $\mathbf{M}$ is the mass matrix. The simulator advances the garment state $\mathbf{q}_n = \left(\mathbf{x}_n,\mathbf{v}_n\right)$ at time step $n$ forward in time at discrete time steps $\Delta t$. $\mathbf{Q}$ consists of states over all time steps $N$.
Although any simulation model can be used, in this work, we make use of XPBD \cite{macklin2016xpbd} due to its excellent performance characteristics. The energy potential $U(\mathbf{x})$ is formulated in terms of a vector of all constraint functions $\mathbf{C}(\mathbf{x})$  and an inverse compliance matrix \(\bm{\alpha}^{-1}\) as
$U(\mathbf{x}) = \frac{1}{2} \mathbf{C}(\mathbf{x})^{\top} \bm{\alpha}^{-1} \mathbf{C}(\mathbf{x}).$ The constraints include triangle constraints, dihedral bending and collision constraints, modelling in-plane stretching and shearing, out-of-plane bending and collisions respectively. At each time step, a position update $\Delta \mathbf{x}$ is computed using a Gauss-Seidel-like iterative solver indexed by $i$ of the following system:
\begin{equation}
\begin{aligned}
    \left(\nabla \mathbf{C}(\mathbf{x}_i)^{\top}\mathbf{M}^{-1}\nabla\mathbf{C}(\mathbf{x}_i) + \tilde{\bm{\alpha}}\right) \Delta \bm{\lambda} &= - \mathbf{C}(\mathbf{x}_i) - \tilde{\bm{\alpha}}\bm{\lambda}_i \\
    \Delta \mathbf{x} &= \mathbf{M}^{-1} \nabla \mathbf{C}(\mathbf{x}_i)\Delta \bm{\lambda},
\end{aligned}
\label{eq:deltaXXpbd}
\end{equation}
where $\tilde{\bm{\alpha}} = \bm{\alpha} / \Delta t^2$ and $\bm{\lambda}$ is the constraint multiplier. Due to the decoupled nature of the solve, the position update $\Delta \mathbf{x}$ can be computed separately for each constraint type. We compute vertex positions as 
\begin{equation}
\begin{aligned}
 \mathbf{x}_{n+1} 
 &= \mathbf{x}_{n} + \Delta \mathbf{x} + \Delta t \left( \mathbf{v}_n + \Delta t \mathbf{M}^{-1} \mathbf{f}_{\text{ext}} \right)
\end{aligned}
\label{eq:xpbdUpdateRule}
\end{equation}
and velocities $\mathbf{v}_{n+1} = \frac{1}{\Delta t} \left( \mathbf{x}_{n+1} - \mathbf{x}_{n}\right)$ where $\mathbf{f}_{\text{ext}}$ denote the external forces acting on the system.

\subsection{Differentiable Cloth Simulation}
\label{subsec:diffsim}
Given a minimizing goal function $\phi$ computed through complex dynamic simulations, differentiable simulation enables gradient-based optimization methods by computing its gradient $\phi$ with respect to the control parameters $\bm{\theta}$ as
\begin{equation}
\frac{d \phi}{d \bm{\theta}} = \frac{\partial \phi}{\partial \mathbf{Q}}\frac{d \mathbf{Q}}{d \bm{\theta}} + \frac{\partial \phi}{\partial \bm{\theta}}
\end{equation}
However, due to the intractability of computing $d \mathbf{Q} / d \bm{\theta}$ directly,  adjoint method is used to replace the vector-matrix product with an equivalent, more efficient computation involving the adjoint of $\mathbf{Q}$, denoted by $\hat{\mathbf{Q}}$ which contains all adjoint states $\hat{\mathbf{q}}_n = \left(\hat{\mathbf{x}}_n \in \mathbb{R}^{3V},\hat{\mathbf{v}}_n \in \mathbb{R}^{3V}\right)$ over all $N$ steps.  We use prior work DiffXPBD \cite{stuyck2023diffxpbd} to compute gradients through the XPBD simulation model. Using the adjoint states, the full derivative $d \phi / d \bm{\theta}$ is obtained using 
\begin{equation}
    \frac{d \phi}{d \bm{\theta}} = \hat{\mathbf{Q}}^{\top} \frac{\partial \Delta \mathbf{x}}{\partial \bm{\theta}} + \frac{\partial \phi}{\partial \bm{\theta}}
    \label{eq:gradientRule}
\end{equation}
where $\Delta \mathbf{x}$ refers to the position updates computed in the XPBD framework in Eq.~\ref{eq:deltaXXpbd}. We refer the readers to \cite{stuyck2023diffxpbd} for detailed derivations of  the adjoint states $\hat{\mathbf{Q}}$  . The quantities $\partial \Delta \mathbf{x} / \partial \bm{\theta}$ and $\partial \phi / \partial \bm{\theta}$ are problem-specific, which we detail in the next sections.

\section{Methodology}

We introduce our computational method for extracting garment and body assets from real 3D scans of clothed humans. Our method uses a differentiable simulator for simultaneous co-optimization of garment 2D pattern shape, cloth material, body pose and shape. See Fig.~\ref{fig:overview} for a visual overview. Starting from an automatically selected template, our goal is to optimize garment patterns and materials that replicate the overall style and fit of the scan. Note that the drape of a given garment, including the wrinkles and surface details, can be different on different body shape and state, and may be adjusted by the wearer, therefore we do not aim to perfectly recreate the garment shapes exactly as they appear in the scan. Additionally, we aim to recover the overall body shape and pose but do not intend to recover other appearance aspects such as the face, since it does not influence the simulated behavior of the clothing.

\subsection{Extracting 3D Garments and Parametric Body}
\label{subsec:dataPreparation}
We process multi-view images to reconstruct and segment a 3D scan and use the resulting geometry to initialize the shape and pose of the parametric body model.
\newline
\noindent \textbf{3D Scan Semantic Garment Segmentation}
From multi-view images of a clothed person, we reconstruct a noisy 3D scan using the 3dMD system \cite{3dmd}. These scans tend to be noisy, contain holes and might not capture regions such as hair or loose clothes accurately. We extract the 3D geometry of the isolated garment(s) of interest using a cloth segmentation algorithm~\cite{fu2019imp} on each of the 18 camera views to obtain per-pixel class predictions. We enforce multi-view class consistency  by selecting the majority garment classes. 
\newline 
\noindent \textbf{Body Shape and Pose Initialization} To fit our parametric body model to the scan, we optimize the body shape $\bm{\nu}$, pose $\bm{\psi}$ and joint lengths by minimizing the Chamfer distance between the vertices and those of the full person scan. We use a Gauss-Newton solver that takes joint limits into account and penalizes self-penetrations of the body mesh.

\subsection{Avatar Optimization}

We leverage differentiable simulation (Sec.~\ref{subsec:diffsim}) to simultaneously recover garment pattern and material, as well as body pose and shape.
Starting from an initial pattern, our method automatically adjusts the size and shape of each panel in the pattern. To achieve this, we require a minimal garment library that defines the pattern structure for each type of garment. We use the semantic information (Sec.~\ref{subsec:dataPreparation}) to automatically identify the garment types. With the estimated body shape and pose, we drape the garment through physical simulation to obtain the initial 3D garment state.

\subsubsection{Optimization Problem Statement}
Once initialized, we aim to find the parameters $\bm{\theta}$ that minimize a loss function $\phi\left( \bm{\theta}, \mathbf{Q} \right)$. The loss function (Sec.~\ref{sec:loss}) encodes how close the geometry is to the segmented scan. The control variables $\bm{\theta}$ include statistical body shape $\bm{\nu}$ and pose $\bm{\psi}$ coefficients to model the body shape under the clothing, material parameters $\bm{\lambda}$ to model the fabric properties and most importantly, the control cage handles $\mathbf{\zeta}$ that deform the 2D pattern space coordinates $\mathbf{p}$ of the garment (Sec.~\ref{sec:controlCage}).

We use gradient descent to optimize these variables over multiple iterations. In each iteration, we run a dynamic differentiable simulation until the garments reach quasi-equilibrium state with the current set of parameters $\bm{\theta}$ to obtain a draped garment. This draped garment is then used to compute a loss, and the gradient information $d \phi / d \bm{\theta}$ is obtained through back-propagation using the differentiable simulator. We determine the full gradient by computing the jacobian $ \partial \Delta \mathbf{x} / \partial \bm{\theta}$ and $\partial \phi / \partial \bm{\theta}$ to evaluate Eq.~\ref{eq:gradientRule}. 
In the following subsections, we explain how to compute gradients with respect to $\bm{\theta}$. 
Note that our pipeline is not limited to the specific implementation of DiffXPBD and can be applied to any differentiable cloth simulation framework.

\subsubsection{Garment Pattern Optimization}
\label{sec:controlCage}

We propose a regularized differentiable cage formulation to effectively and robustly optimize for the 2D patterns of garments such that the simulated and draped 3D representation of the garment closely aligns with the scan.

\noindent\textbf{Control Cage Pattern Representation}:
 The 3D positions of each garment is controled by its corresponding 2D pattern (Fig.~\ref{fig:patternspaces}). While it is possible to directly optimize for the 2D pattern vertices $\mathbf{p}$ directly, this approach is highly non-regularized and can produce ill-shaped or even non-physical inverted rest shape geometries that cause simulators to fail. 
\begin{wrapfigure}{r}{0.13\textwidth}
\vspace{-10pt}
\hspace{-27pt}
\centering
\includegraphics[width=0.18\textwidth]{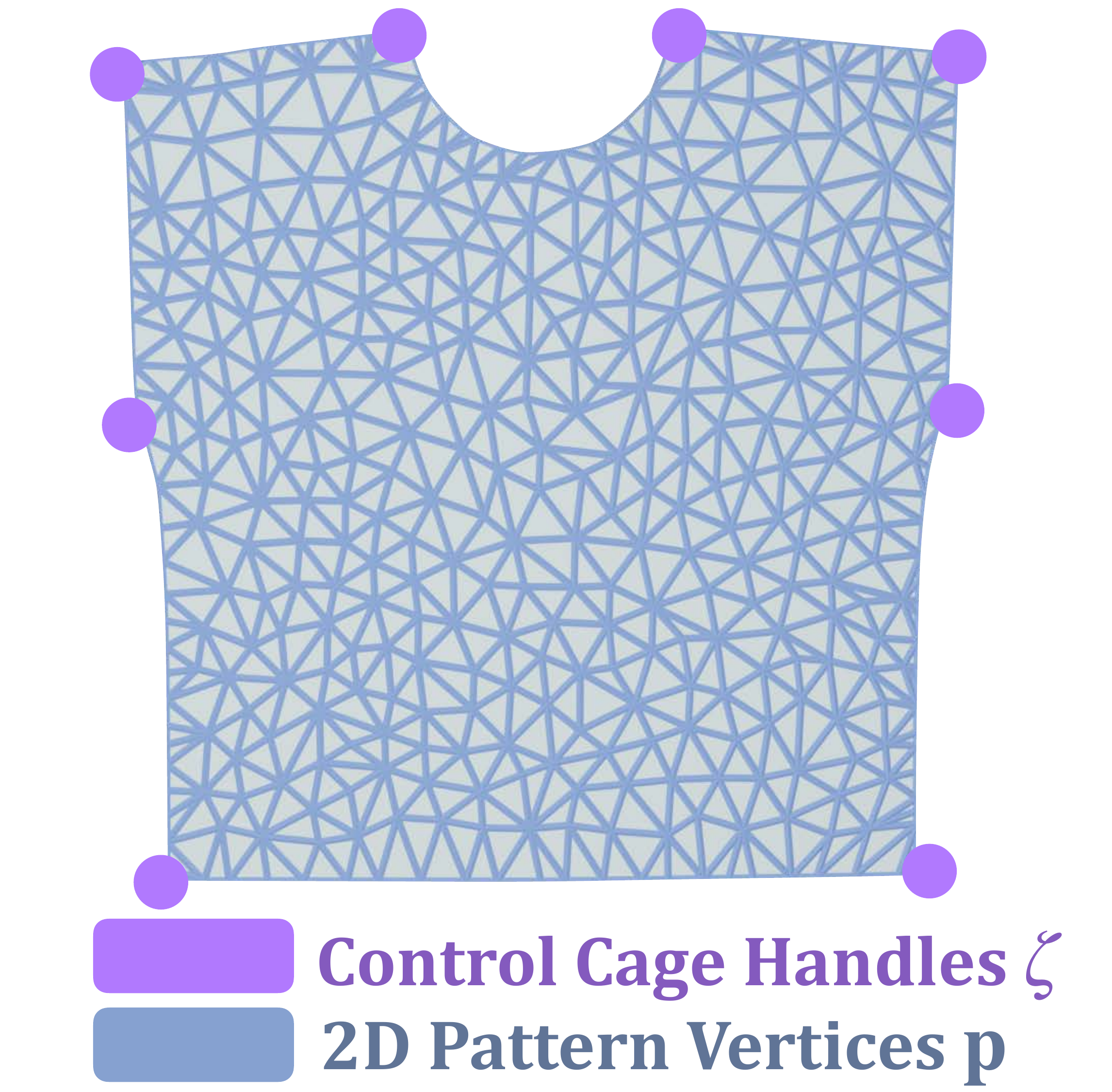}
\vspace{-25pt}
\end{wrapfigure}%
A high number of optimization variables can also cause the optimization to get stuck in a local minimum  (See our ablation study in Sec.~\ref{sec:ablation}). Additionally, directly optimizing for the 2D coordinates does not respect design constraints that are better represented in a limited subspace of reasonable designs. Therefore, we further regulate the optimization problem by selecting and optimizing a set of 2D control vertices $\zeta$ on the boundaries of the individual panels of the 2D pattern that directly deform and manipulate the underlying 2D patterns through control cages instead. 

\noindent\textbf{Control Cage Handle Selection}: We use the geometric information of the 2D garment patterns to automatically identify control cage points, see the inset figure above. Our algorithm first extracts the boundary loop of the underlying mesh for each connected component representing a garment panel in the 2D garment pattern, then processes the boundary loop and marks a vertex as a control point if it lies on the convex hull of the pattern or when its local curvature exceeds a threshold (10\textdegree~in our implementation).

\noindent\textbf{Differentiable Control Cage Optimization}: We use the control handles to deform the underlying 2D pattern via Mean Value Coordinates~\cite{ju2005mean}. During initialization, we computes a generalized barycentric coordinate for each vertex in the 2D pattern with respect to each vertex on the control cage, expressed as $\Bar{\mathbf{x}} = \bm{W} \bm{\zeta}$. We compute the required derivatives to evaluate Eq.~\ref{eq:gradientRule} following the chain rule as:
\begin{equation}
\frac{\partial \Delta \mathbf{x}}{\partial \bm{\zeta}} = \frac{\partial \Delta \mathbf{x}}{\partial \Bar{\mathbf{x}}}  \frac{\partial\Bar{\mathbf{x}}}{\partial \bm{\zeta}} = \frac{\partial \Delta \mathbf{x}}{\partial \Bar{\mathbf{x}}} \bm{W}
\end{equation}

We detail the derivation for $\frac{\partial \Delta \mathbf{x}}{\partial \Bar{\mathbf{x}}}$ in the Supplementary Material.
\vspace{-0.2cm}
\subsubsection{Body Shape and Pose Optimization}
\label{subsec:bodyShapeEstimation}

The initial body shape obtained from the geometry-based optimization (Sec.~\ref{subsec:dataPreparation}) only uses the geometric information from the scan. 
We improve accuracy for the body shape and pose through differentiable simulation to explicitly account for the separate cloth geometry layer on top of the body.
The body interacts with the garments during simulation solely through the collisions. Therefore we only need to compute the derivatives for the collision response updates $\partial \Delta \mathbf{x}_\text{cloth-body collision}$ to evaluate Eq.~\ref{eq:gradientRule}. Using chain rule, we compute
\begin{equation}
\frac{\partial \Delta \mathbf{x}_\text{cloth-body collision}}{\partial \bm{\alpha}} = 
\frac{\partial \Delta \mathbf{x}_\text{cloth-body collision}}{\partial \mathbf{x}_\text{body} } \frac{\partial \mathbf{x}_\text{body} }{\partial \bm{\alpha}}
\end{equation}
where $\bm{\alpha}$ is either body shape $\bm{\nu}$ or pose $\bm{\psi}$. The first term $\partial \Delta \mathbf{x}_\text{cloth-body collision} / \partial \mathbf{x}_\text{body}$ measures how the cloth position updates changes with a change in position update of the body vertices. For the body shape parameters, the final term is computed by back-propagating the gradients through the body shape model described in Sec.~\ref{eq:PCA}. To obtain joint angle gradients, we differentiate through the linear blend skinning operation.
 
\subsubsection{Material Property Estimation}

We optimize for cloth material properties to better match the shape of the scanned garments. \rev{See the supp. material for details.} The bending parameter has the most significant effect \cite{wang2022learnbend} on the wrinkling of the cloth and can be inferred from draped garments. Since the bending material parameter $\lambda$ only enters the computational graph when computing the dihedral constraint, computing the jacobian reduces to $\partial \Delta \mathbf{x} / \partial \lambda = \partial \Delta \mathbf{x}_\text{Dihedral} / \partial \lambda$.

\vspace{-0.2cm}
\subsubsection{Loss Function Design}
\label{sec:loss}
Our loss function is designed with two main components: the \emph{feature matching} term $\mathcal{L}_\text{features}$ and the \emph{regularization} term $\mathcal{L}_\text{regularization}$. The feature matching term encourages the optimization to converge to the scan in the 3D world coordinates after simulations, while the regularization terms operate on the 2D patterns to maintain desired features. The loss function is thus given by $\phi = \mathcal{L}_\text{features} + \mathcal{L}_\text{regularization}$.

\paragraph{Feature Matching} is used to ensure that the simulated garment matches the scan. We use two distinct terms with individual weights $\rho$ and $\sigma$ to achieve this goal. The \emph{boundary loss} term measures how well the boundaries overlap and serves to drive correct lengths of the pattern to match size, whereas the \emph{interior loss} is designed to match the looseness of the fit: $\mathcal{L}_\text{features} = \rho \mathcal{L}_\text{boundary} + \sigma \mathcal{L}_\text{interior}$.

\noindent\textbf{Boundary Feature Matching} We segment boundary points on the scan and align and match with those of the simulated garment and minimize the L2 distance to boundaries such as sleeve lengths and hems.

\noindent\textbf{Interior Point Feature Matching} We measure and minimize the Chamfer distance between the interior points of the simulated garment and target segmented garment scan.

\paragraph{Regularization}
To improve our loss formulation, we include regularization terms that act on the pattern space to maintain desired features in the design. We enable the optimizer to adapt individual patterns in the garment design to match features in the 3D scan. However, this process could result in designs where seams that are to be sewn together have different edge lengths, leading to undesired artifacts such as gathering, which produces a ruffled effect. To prevent this, we add two regularizers with weights $\alpha$ and $\beta$, giving
$\mathcal{L}_\text{regularization} = \alpha \mathcal{L}_\text{seam length} + \beta \mathcal{L}_\text{curvature}$.
We penalize seam length differences for edges on the individual patterns that are to be sewn together, as shown in Fig.~\ref{fig:patternspaces}. The color coded seams that are to be sewn together should have the same length. We color-coded a subset of the seams since the right half is symmetric. Mathematically, we express this as follows
\begin{equation}
\mathcal{L}_\text{seam length} = \sum_{i \in \text{Seam edges}} || \mathbf{p}_{i} - \mathbf{p}_{i+1} ||^2 - || \mathbf{p}'_i - \mathbf{p}'_{i+1}||^2
\end{equation}

Additionally, to prevent noisy and undesired designs, we penalize the changes in boundary curvature of the 2D pattern with respect to the original garment template similar to the work of Wang~\cite{WangRuleFreeSewing2018}. We seek a scaled rotation matrix $\mathbf{T}_i = s \mathbf{R}_i \in \mathbb{R}^{2 \times 2 }$ at each point $\mathbf{p}_i$ with least curvature distortion to its connected boundary edges, $\mathbf{T}_i = \arg \min_{\mathbf{T}} || \mathbf{e}_{i1} - \mathbf{T} \Bar{\mathbf{e}}_{i1} ||^2 + || \mathbf{e}_{i2} - \mathbf{T} \Bar{\mathbf{e}}_{i2} ||^2$ with $\mathbf{e}_{i1} = \mathbf{p}_{i + 1} - \mathbf{p}_i$ and  $\mathbf{e}_{i2} = \mathbf{p}_{i - 1} - \mathbf{p}_i$. The loss is defined as the accumulation of the curvature distortion as 
\begin{equation}
\mathcal{L}_\text{curvature} = \sum_{i \in \partial\Omega} w_i || (\mathbf{e}_{i1} - \mathbf{T}_i \Bar{\mathbf{e}}_{i1}) + (\mathbf{e}_{i2} - \mathbf{T}_i \Bar{\mathbf{e}}_{i2}) ||^2,
\end{equation}
where the quantities denoted by $\Bar{\cdot}$ refer to the UV coordinates in the original garment pattern.


\section{Experiments}
\noindent\textbf{Data}: We evaluate our method on a variety of 3D scans of humans captured with a 3dMD\cite{3dmd} system. We select 4 subjects wearing different garments (dress, long-sleeve, polo, shirt) and obtain their corresponding 3D reconstructions. Note that these scans tend to be noisy and contain holes. Nevertheless, they can still serve as 3D targets during the optimization process.
Since there are no perfect or clean ``ground-truth'' 3D scans available for evaluation, we have asked a skilled professional artist to create virtual garments that match the scans to the best of their ability. This process serves as our upper quality bar. Therefore, we provide quantitative comparisons using both the 3D scan input and the artist-made garments as ground-truth to demonstrate the clear improvements \papername~offers over previous work.

\noindent\textbf{Baselines}: We evaluate the output geometry of our method against two methods that employ 2D images as an input and one that uses 3D scans. Specifically, we run PiFU-HD~\cite{saito2020pifuhd} on the clothed human scan and segment the garment in 3D to use it for evaluations. We also employed a recent diffusion-based approach that given a single garment image, synthesizes six multi-view consistent novel views, we then use NeuS\cite{wang2021neus} to extract the 3D geometry. We also compare our method against PoP\cite{ma2021power} which is the closest to our work as it uses 3D scans as an input and outputs a point cloud of the clothed human which we pass to Poisson reconstruction to obtain a mesh. Finally, we provide a comparison of the 2D patterns against NeuralTailor~\cite{Korosteleva2022} which estimates 2D garment patterns from 3D point clouds of a draped garment in \figref{fig:patternComparison}.

\noindent\textbf{Evaluation Metrics}: Similar to ~\cite{su2023caphy} we use the Chamfer Distance (CD) for comparisons directly in 3D, and LPIPS\cite{zhang2018perceptual} and SSIM\cite{wang2004image} for perceptual metrics in the 2D space using the exact same rendering conditions for all methods. We evaluate the mesh quality of the results using the triangle conditioning metric~\cite{shewchuk2002quality}. All metrics are reported in Tab.~\ref{tab:CDMeasurements}.

\noindent\textbf{Implementation Details}: Our method is implemented in C++ using open-source libraries such as Eigen and libigl. All experiments were conducted on a machine with 14-core i7 CPU with 32GB RAM. Note that there is no limitation to implement our method on GPU, and it would benefit from our implementation since the expensive Jacobian computations map well to highly parallel GPU code. The linear solve can be further accelerated with cuSPARSE. 

\noindent \textbf{Performance}: \papername~incorporates simulation within an optimization process to yield high-quality outcomes at the expense of increased computational requirements. Optimization takes about one minute per iteration and total times vary between 20 to 200 minutes depending on the garment and iteration count. Baseline methods range between several seconds to 10 minutes for complete inference results. Note that our method can be executed in batch mode automatically. Our method runs on CPU, whereas the baselines are run on an NVIDIA GeForce RTX 4080 GPU.

\subsection{Garment Pattern Optimization}
\begin{figure}
    \centering
    \includegraphics[width=0.99\columnwidth, trim={0 0 0 0},clip]{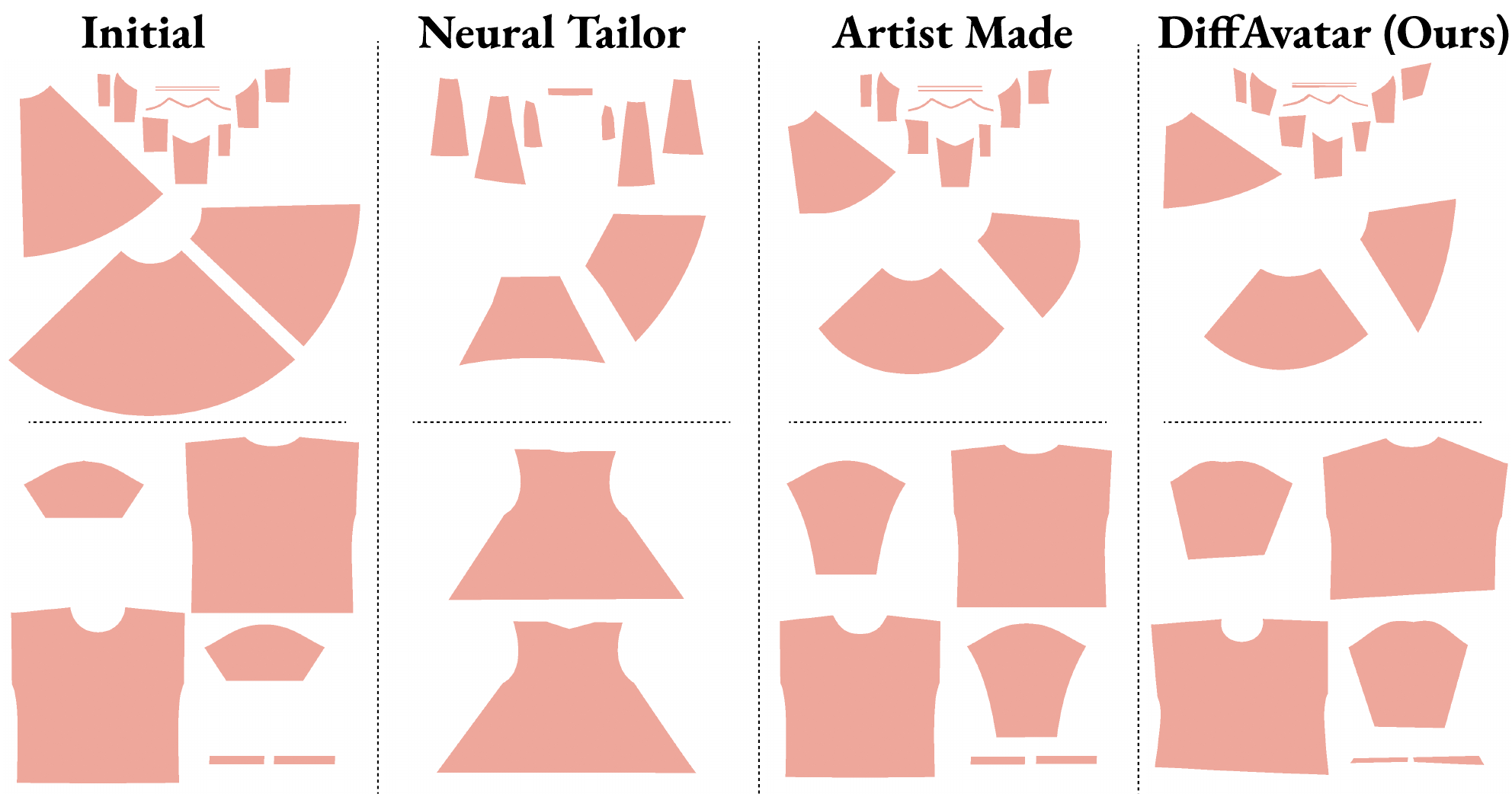}
    \caption{\textbf{2D pattern comparison.} The automatically optimized 2D patterns of the dress (first row) and long sleeve shirt (second row) by \papername{} closely match the manually created artist ones. However, those generated by NeuralTailor~\cite{Korosteleva2022} do not resemble the artist-made patterns closely and miss important details.}
    \label{fig:patternComparison}
    \vspace{-0.15cm}
\end{figure}

The corresponding 2D optimized patterns for the dress and long sleeve shirt are shown in \figref{fig:patternComparison}.  Starting from the initial panel in the first column, \papername{} (last column) generates patterns closely resembling the one designed manually by an artist (second last column). In contrast, although NeuralTailor~\cite{Korosteleva2022} (second column) does not need an initial template, the result can be far from the target and can even be missing key features like the shirt sleeves.

\subsection{Body and Cloth Material Optimization}

\begin{figure}
    \centering
    \includegraphics[width=0.45\textwidth, trim={0 0 0 0},clip]{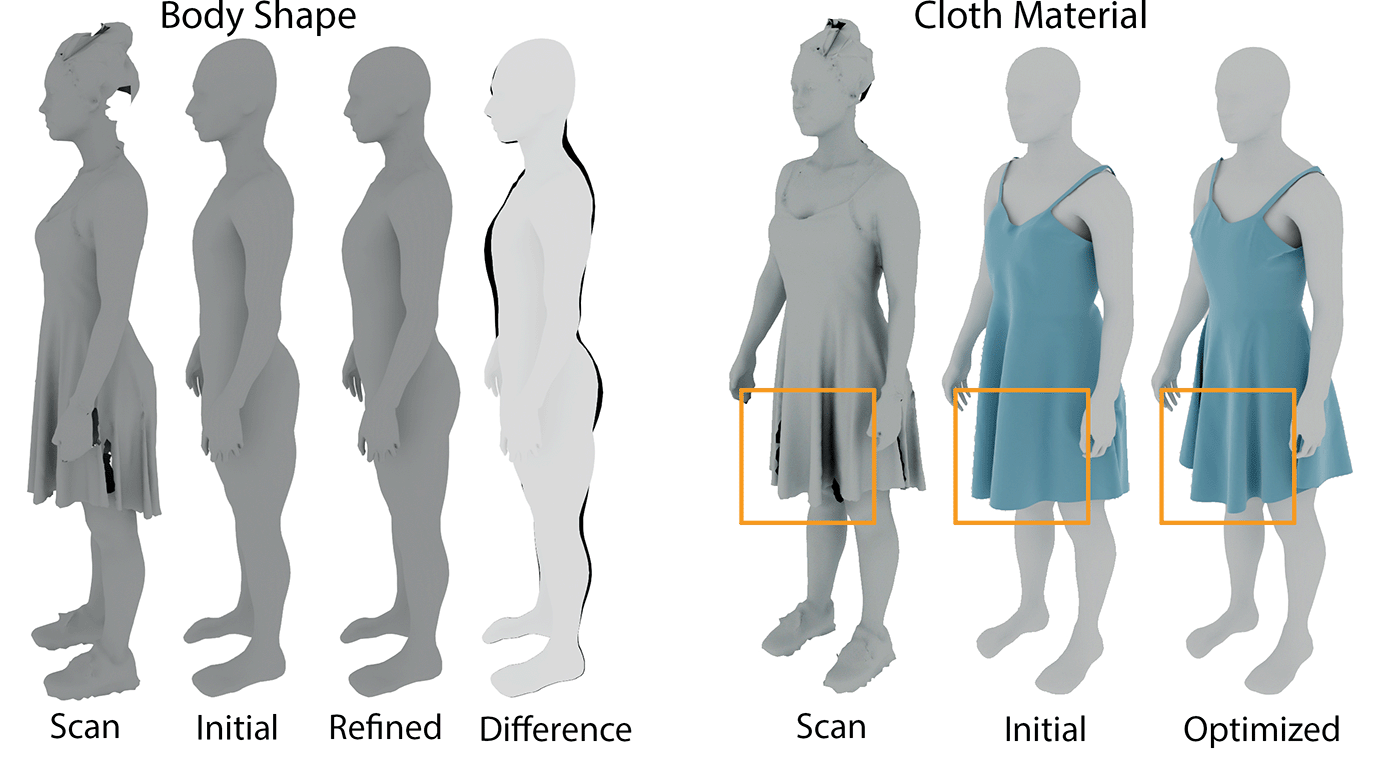}
    \caption{\textbf{Body shape and cloth material estimation}. \textit{Left}: We fit a statistical body model to the 3D scan and refine this estimate using our differentiable simulation pipeline and show the difference in shape between initial and refined in black. 
    \textit{Right}: Our initial material estimate produces large folds that do not match the scan as well as our optimized result shown rightmost.}
    \label{fig:bodyShapeOptimization}
\end{figure}

We visualize different stages of our body shape estimation in \figref{fig:bodyShapeOptimization} (left) and demonstrate how our physics-aware method improves the estimated body shape. The right of the figure demonstrates our ability to recover cloth material properties to closely match the garment drape in the scan.

\begin{table}[t]
    \centering
    \setlength{\tabcolsep}{0.5mm}
    \renewcommand{\arraystretch}{1.2}
    \resizebox{\columnwidth}{!}{
    \begin{tabular}{lccccccc}
        \toprule
        \multirow{2}{*}{Method} &  \multicolumn{3}{c}{Against GT Scan} &  \multicolumn{3}{c}{Against GT Artist} & \multirow{2}{*}{Mesh} \\   \cmidrule(l){2-4} \cmidrule(l){5-7} 
        ~ & \multirow{2}{*}{CD $\downarrow$}  & \multirow{2}{*}{LPIPS $\downarrow$} & \multirow{2}{*}{SSIM $\uparrow$}  & \multirow{2}{*}{CD $\downarrow$}  &  \multirow{2}{*}{LPIPS $\downarrow$} & \multirow{2}{*}{SSIM $\uparrow$} & Quality $\uparrow$ \\
        & & & & & & & min(avg) \\ \hline
        
        Scan & - & - & - & 1.045  &  0.127 & 0.852& 0.099(0.422) \\ 
        Artist & 1.045 & 0.127 & 0.852  & -  & -  & -& 0.171(0.389) \\
        Initial & 3.071 & 0.165 & 0.815 & 3.396 & 0.123 & 0.859 & 0.188(0.391) \\
        \hline
        PiFU-HD~\cite{saito2020pifuhd} & 1.930 & 0.145 & 0.836 & 2.009 & 0.129 & 0.836& 0.000(0.305) \\
        \small{Diffusion+NeuS}~\cite{wang2021neus} & 3.410 & 0.171 & 0.799 & 3.362 & 0.177 & 0.797 & 0.000(0.266) \\
        PoP~\cite{ma2021power} & 1.695 & 0.140 & 0.831 & 1.866 & 0.092 & 0.842 & 0.000(0.316) \\
        \hline
        \textbf{DiffAvatar (Ours)} & \textbf{1.311} & \textbf{0.133} &\textbf{0.842} & \textbf{1.688} & \textbf{0.085} & \textbf{0.893} & {\bf 0.143}({\bf 0.373}) \\
      \bottomrule
    \end{tabular}}
    \caption{\textbf{Quantitative Comparisons}. 
    We used the 3D scan and artist-made mesh as ground truth to evaluate our method on the dress example. Our results show that we achieve the closest CD fit, best perceptual metrics, and produce good mesh quality. In contrast, all competing methods produce a minimum mesh quality of 0 (or near 0), making their output unsuitable for simulation.}
    \label{tab:CDMeasurements}
\end{table}

\subsection{Method Evaluations}
\begin{figure*}
    \centering
    \includegraphics[width=0.99\textwidth, trim={0 0 0 0}]{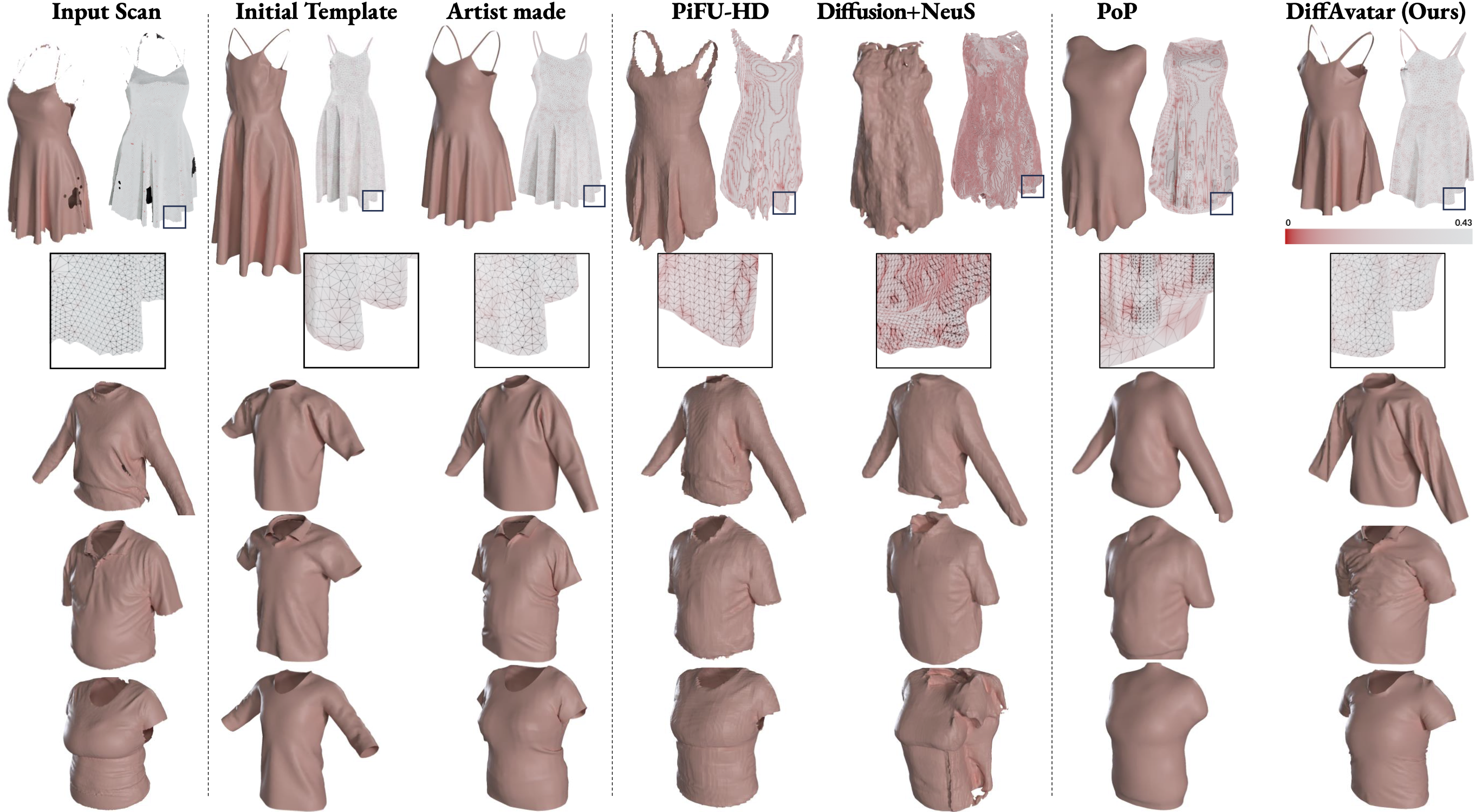}
    \caption{\textbf{Qualitative Comparisons.} \papername~faithfully captures garments with natural draping behavior and wrinkle details where all prior works fail to reconstruct simulation-ready meshes. \textbf{Mesh quality} (\textit{Top row}). The generated mesh quality of the mesh visualized with red-to-white gradient representing lowest to highest quality. \papername~generates simulation-ready meshes of high quality, comparable to artist-made meshes where 2D prior works such as PiFU-HD~\cite{saito2020pifuhd} and Diffusion+NeuS~\cite{wang2021neus}, or 3D works such as PoP~\cite{ma2021power} come short.}
    \label{fig:methodComparisons}
    \vspace{-0.1cm}
\end{figure*}

We evaluate the reconstructed 3D geometry of our approach against three prior works (\figref{fig:methodComparisons}) and report quantitative results in \tabref{tab:CDMeasurements}. Regardless of the ground-truth considered (scan or artist-made), \papername~outperforms all prior works across both 2D and 3D metrics. PiFU-HD~\cite{saito2020pifuhd} and Diffusion+NeuS reconstruct the frontal part of the geometry fairly well but the back side is smooth and all results lack fine-level details (wrinkles and folds). 
PoP works better for tight-fit garments but fails to accurately reconstruct the details of a dress, producing closed-surface meshes without arm holes that are unsuitable for simulation. It is evident from these results
that our approach is the only one which faithfully captures the garment with simulation-ready topology.
To evaluate mesh quality, we apply a conditioning quality metric~\cite{shewchuk2002quality}
to the 3D meshes for a fair comparison with the baseline methods that do not produce rest shape geometry.
Prior methods produce a near 0 minimum mesh quality, which indicates the presence of poorly-conditioned or zero-area triangles that are unsuitable for simulation. Our results show favorable quality compared to all past works.

\subsection{Ablation Studies}\label{sec:ablation}
\begin{figure}
    \centering
    \includegraphics[width=0.49\textwidth, trim={0 0 0 0},clip]{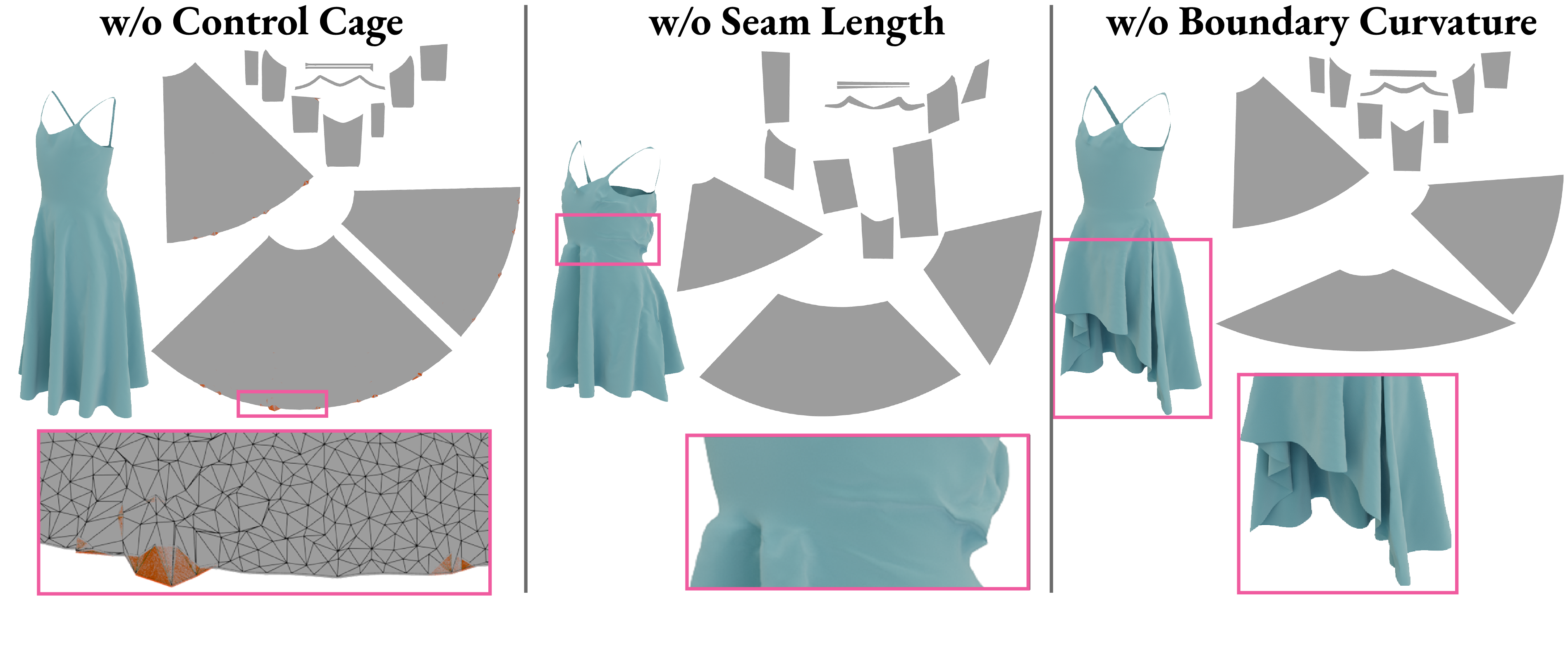}
    \caption{\textbf{Ablation Study}. \textit{Left}: w/o control cage, the optimizer quickly produces inverted non-physical triangle elements (highlighted in orange) in the rest shape which causes any simulator to fail. \textit{Middle}: w/o seam length regularization, the seam lines do not match leading to excessive amount of fabric. \textit{Right}: w/o boundary curvature regularization, the pattern distorts into unwanted shapes.}
    \label{fig:ablation}
    \vspace{-0.15cm}
\end{figure}

We perform an ablation study on the importance of individual components in our design, see~\figref{fig:ablation} and \tabref{tab:Ablations}. We demonstrate that our control cage formulation is crucial for producing physically correct results that can be simulated. The seam length regularization is necessary to prevent seam length mismatches, which leads to excessive gathering of the fabric. The boundary curvature regularization is required to preserve the design intent of the garment.

\subsection{Novel Simulated Sequences}
In contrast to baseline methods, \papername~is the only one that generates high-quality simulation-ready geometry with associated 2D rest shape and cloth material properties enabling us to create new simulations that are faithful to the original garment with ease. \figref{fig:teaser} shows select frames from novel simulated sequences using the optimized dress. See the supplemental material for additional animations.

\begin{table}[t]
    \centering
    \setlength{\tabcolsep}{0.85mm}
    \renewcommand{\arraystretch}{1.2}
    \begin{tabular}{lccc}
    \toprule
        Method Variant & CD $\downarrow$  & LPIPS $\downarrow$ & SSIM $\uparrow$  \\ \hline
        w/o Control Cage & 3.866 & 0.168 & 0.792  \\ 
        w/o Seam Length Term & 1.409 & 0.115 & 0.843 \\ 
        w/o Boundary Curvature Term & 2.249 & 0.124 & 0.838  \\ 
        DiffAvatar (Complete) & \textbf{1.122} & \textbf{0.102} & \textbf{0.863}  \\ 
        \bottomrule
    \end{tabular}
    \caption{\textbf{Ablation Study.} By removing each of the proposed components of \papername~we showcase their impact in 3D with Chamfer Distance and 2D with perceptual metrics to the final result for the \textit{dress} against the ground-truth scan.}
    \label{tab:Ablations}
    \vspace{-0.3cm}
\end{table}

\subsection{Limitations and Future Work}
Our method optimizes through a continuum of pattern variations starting from a template based on the garment category. Although we do not address discrete changes in the number of pattern pieces or mesh topology, such a system can be incorporated into the pipeline retroactively. 
We use dynamic simulation but rely on states close to quasi-equilibrium. This implies that draped garments with strong friction or dynamic effects can be challenging to estimate and can produce a different final aesthetic. 
Since we are already using a dynamic simulator, a straightforward extension is to match dynamic sequences and recover additional parameters.
For multi-layered clothing, garments can be occluded, making recovery a fundamentally difficult goal. Our method is well suited to handle occlusions due to its strong physical priors about the behavior fabric. 
\section{Conclusion}
We introduced \papername~a new approach that utilizes differentiable simulation for scene recovery to generate high-quality, physically plausible assets that can be used for simulation applications. 
Our method considers the complex non-linear behavior of cloth and its intricate interaction with the underlying body when optimizing for scene parameters in a unified and coupled manner that takes into account the interplay of all components. 
We showcased that \papername~outperforms prior works across different metrics, producing high-quality garment results in both 3D and the 2D pattern space and generates simulation-ready assets close to those that are manually designed by a trained artist.


{\small
\bibliographystyle{ieee_fullname}
\bibliography{References}
}

\clearpage
\appendix
 \maketitle

 \newcommand{\xbar}{\Bar{\mathbf{x}}}

\twocolumn[{%
\renewcommand\twocolumn[1][]{#1}%
\maketitle
\vspace{-0.65cm}
\centering
    \includegraphics[width=0.99\textwidth]{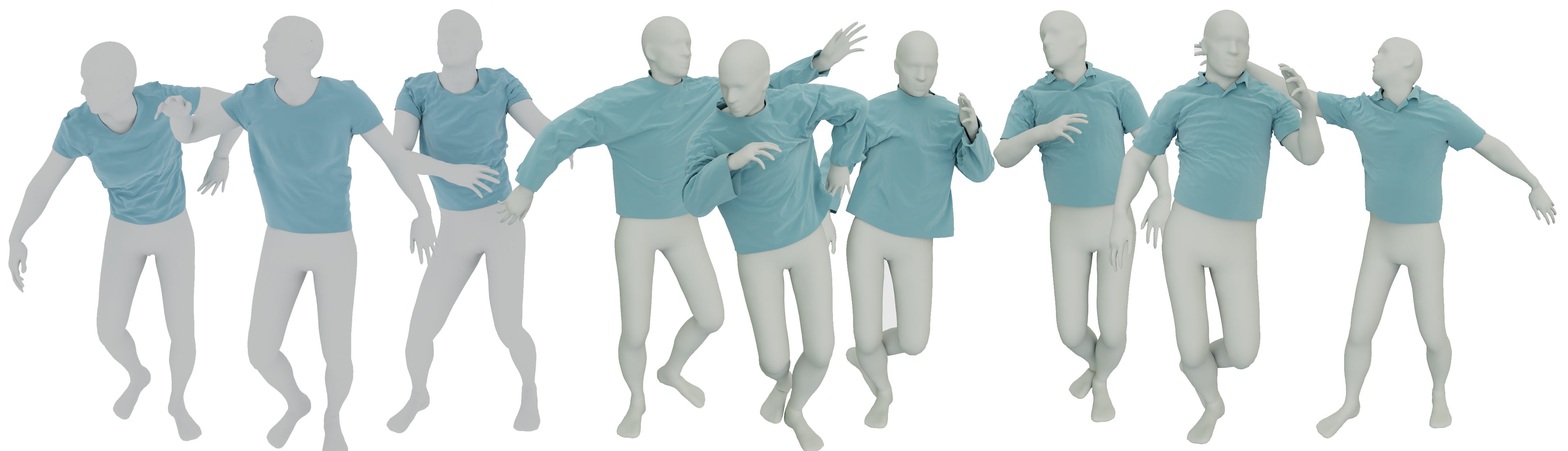}
    \captionof{figure}{After recovering simulation-ready assets, we can easily generate novel simulation results.}
    \vspace*{0.2cm}
    \label{fig:shirtDance}
}]

\section{DiffXPBD: Differentiable Simulation}


We provide details on the implementation of our differentiable simulator which builds upon DiffXPBD \cite{stuyck2023diffxpbd}. The simulation moves the states forward in time using $\mathbf{q}_{n+1} = \mathbf{F}_n\left(\mathbf{q}_{n+1}, \mathbf{q}_n, \mathbf{u}\right)$, see Eq. \eqref{eq:xpbdUpdateRule}. The adjoint states $\hat{\mathbf{Q}}$ are computed in a backward pass using \begin{equation}
    \hat{\mathbf{q}}_{n-1} =  \left( \frac{\partial \mathbf{F}_{n-1}}{\partial \mathbf{q}_n}  \right)^\top \hat{\mathbf{q}}_{n-1} + \left( \frac{\partial \mathbf{F}_n}{\partial \mathbf{q}_n} \right)^\top \hat{\mathbf{q}}_n + \left( \frac{\partial \phi}{\partial \mathbf{q}_n} \right)^{\top}
    \label{eq:adjointRule}
\end{equation}

The XPBD simulation frameworks uses the following update scheme.
\begin{equation}
\begin{aligned}
 \mathbf{x}_{n+1} 
 &= \mathbf{x}_{n} + \Delta \mathbf{x}\left(\mathbf{x}_{n+1} \right) + \Delta t \left( \mathbf{v}_n + \Delta t \mathbf{M}^{-1} \mathbf{f}_{\text{ext}} \right) \\
 \mathbf{v}_{n+1} 
 &= \frac{1}{\Delta t} \left( \mathbf{x}_{n+1} - \mathbf{x}_{n} \right)
\end{aligned}
\label{eq:xpbdUpdateRule}
\end{equation}

We find the adjoint evolution for the XPBD integration scheme by combining this with \eqref{eq:adjointRule} as
\begin{equation}
\begin{aligned}
    \hat{\mathbf{x}}_{n} 
    &= \hat{\mathbf{x}}_{n+1} + \left(\frac{\partial \Delta \mathbf{x}}{\partial \mathbf{x}} 
    + \Delta t^2 \mathbf{M}^{-1} \frac{\partial \mathbf{f}_{\text{ext}}}{\partial \mathbf{x}} \right)^\top \hat{\mathbf{x}}_{n} \\
    &+ \frac{\hat{\mathbf{v}}_n}{\Delta t}
    - \frac{\hat{\mathbf{v}}_{n+1}}{\Delta t}
    + \frac{\partial \phi}{\partial \mathbf{x}}^\top \\
    \hat{\mathbf{v}}_n 
    &= \left( \frac{\partial \Delta \mathbf{x}}{\partial \mathbf{v}} + \Delta t^2 \mathbf{M}^{-1} \frac{\partial \mathbf{f}_{\text{ext}}}{\partial \mathbf{v}} \right)^\top \hat{\mathbf{x}}_{n} \\
    &+ \Delta t \hat{\mathbf{x}}_{n+1} + \frac{\partial \phi}{\partial \mathbf{v}}^\top
\end{aligned}
\label{eq:rawAdjoint}
\end{equation}

After re-arranging and by substituting $\hat{\mathbf{v}}_n$ we find the adjoint states as

\begin{equation}
\begin{aligned}
&\left( I - \frac{\partial \Delta \mathbf{x}}{\partial \mathbf{x}} - \Delta t^2 \mathbf{M}^{-1} \frac{\partial \mathbf{f}_{\text{ext}}}{\partial \mathbf{x}} - \frac{1}{\Delta t}\frac{\partial \Delta \mathbf{x}}{\partial \mathbf{v}} - \Delta t \mathbf{M}^{-1}\frac{\partial \mathbf{f}_{\text{ext}}}{\partial \mathbf{v}} \right)^\top \hat{\mathbf{x}}_n \\ 
&= 2\hat{\mathbf{x}}_{n+1} - \frac{\hat{\mathbf{v}}_{n+1}}{\Delta t}  + \frac{\partial \phi}{\partial \mathbf{x}}^\top + \frac{1}{\Delta t} \frac{\partial \phi}{\partial \mathbf{v}}^\top
\end{aligned}
\end{equation}
\label{eq:adjointXPBD}

\subsection{Material Model}
We use the orthotropic StVK model for modeling stretching and shearing and a hinge-based bending energy as detailed in \cite{stuyck2023diffxpbd}. The material parameters are recovered as part of the optimization process. Different material models can also be used.

\section{Gradient of 3D Cloth Positions to 2D Patterns}
To compute the gradient of the position with respect to the 2D patterns, we need to compute $\frac{\partial\Delta \mathbf{x}}{\partial \xbar_i}$, for each of the 2D cloth vertex i $\in [0, 1,\dots, n]$.  We use the same set of constraints as in DiffXPBD, where $\mathbf{C} = \left[\bm{\epsilon}_{00}, \bm{\epsilon}_{11}, \bm{\epsilon}_{01} \right]$, and $\bm{\epsilon}$ is the Green strain. 

\begin{equation}
\begin{aligned}
\frac{\partial \Delta \mathbf{x}}{\partial \xbar_i} &= \mathbf{M}^{-1} \left( \frac{\partial \nabla \mathbf{C}}{\partial \xbar_i} \Delta \bm{\lambda} + \nabla \mathbf{C} \frac{\partial \Delta \bm{\lambda}}{\partial \xbar_i} \right)
\end{aligned}
\end{equation}

\begin{equation}
\begin{aligned}
\frac{\partial \Delta \bm{\lambda}}{\partial \xbar_i} 
\hspace{0.2cm} &= -\mathbf{J}^{-1} \frac{\partial \mathbf{J}}{\partial \xbar_i} \mathbf{J}^{-1} \mathbf{b} + \mathbf{J}^{-1}\frac{\partial \mathbf{b}}{\partial \xbar_i} \\
\hspace{0.2cm} &= -\mathbf{J}^{-1} \left(\frac{\partial \mathbf{J}}{\partial \xbar_i} \Delta \bm{\lambda} - \frac{\partial \mathbf{b}}{\partial \xbar_i}\right)           
\end{aligned}
\end{equation}
where $\frac{\partial \mathbf{b}}{\partial \xbar_i}$ and $\frac{\partial \mathbf{J}}{\partial \xbar_i}\Delta \bm{\lambda}$ are computed as
\begin{equation}
\begin{aligned}
\frac{\partial \mathbf{b}}{\partial \xbar_i} 
\hspace{0.2cm} &= -\frac{\partial \mathbf{C}}{\partial \xbar_i} - \frac{\partial \tilde{\bm{\alpha}}}{\partial \xbar_i}\bm{\lambda} - \tilde{\bm{\alpha}} \frac{\partial \bm{\lambda}}{\partial \xbar_i} \\
\hspace{0.2cm} &= - \nabla \mathbf{C} - \tilde{\bm{\alpha}} \sum \frac{\partial \Delta \bm{\lambda}}{\partial \xbar_i}
\end{aligned}
\end{equation}
\begin{equation}
\begin{aligned}
\frac{\partial \mathbf{J}}{\partial \xbar_i}\Delta \bm{\lambda} 
&= \frac{\partial \nabla \mathbf{C}^T}{\partial \xbar_i} \mathbf{M}^{-1} \nabla \mathbf{C} \Delta \bm{\lambda} + \nabla \mathbf{C}^T \mathbf{M}^{-1} \frac{\partial \nabla \mathbf{C}}{\partial \xbar_i}\Delta \bm{\lambda} \\
&= \frac{\partial \nabla \mathbf{C}^T}{\partial \xbar_i} \Delta \mathbf{x} + \nabla \mathbf{C}^T \mathbf{M}^{-1} \frac{\partial \nabla \mathbf{C}}{\partial \xbar_i}\Delta \bm{\lambda}
\end{aligned}
\end{equation}

Given that  $\bm{\epsilon}$ is a function of the deformation gradient $\mathbf{F}$, we provide the gradient of $\mathbf{F}$ with respect to the rest positions, and the rest should just follow from chain rule. Note that $\mathbf{F} = \mathbf{D}\Bar{\mathbf{D}}^{-1}$, where the columns of $\mathbf{D}$ and $\Bar{\mathbf{D}}$ are the edge vectors, such that

\begin{equation}
\begin{aligned}
\mathbf{D} &= 
\begin{bmatrix}
\mathbf{x}_0-\mathbf{x}_2 & \mathbf{x}_1-\mathbf{x}_2
\end{bmatrix} \\
\Bar{\mathbf{D}} &= 
\begin{bmatrix}
\xbar_0-\xbar_2 & \xbar_1-\xbar_2
\end{bmatrix}
\end{aligned}
\end{equation}
The dimensions of the matrix $\mathbf{D}$ is 3x2, $\Bar{\mathbf{D}}$ is 2x2, and $\mathbf{F}$ is 3x2. 

We compute the derivative of the deformation gradient using the Einstein notation for $\xbar_0$ and $\xbar_1$
\begin{equation}
\begin{aligned}
        \frac{\partial\mathbf{F}_{ij}}{\partial\xbar_{mn}} &= \mathbf{D}_{ik}\frac{\partial\bar{\mathbf{D}}_{kj}^{-1}}{\xbar_{mn}} \\
        &= - \mathbf{D}_{ik} \bar{\mathbf{D}}^{-1}_{k\alpha} \frac{\partial\bar{\mathbf{D}_{\alpha\beta}}}{\partial\xbar_{mn}}\bar{\mathbf{D}_{\beta j}^{-1}}\\
        &= -\mathbf{D}_{ik} \bar{\mathbf{D}}^{-1}_{km}\bar{\mathbf{D}}^{-1}_{nj},
\end{aligned}
\end{equation}
where $\xbar_{mn}$ is the nth component of $\xbar_m$.
\begin{equation}
     \frac{\partial\mathbf{F}_{ij}}{\partial\xbar_2} = - (\frac{\partial\mathbf{F}_{ij}}{\partial\xbar_0} + \frac{\partial\mathbf{F}_{ij}}{\partial\xbar_1})
\end{equation}

\section{Novel Animations}

Figure~\ref{fig:shirtDance} shows select frames from a novel simulated sequence with the recovered body shapes and garment patterns and materials.

\end{document}